\documentclass[runningheads]{llncs}
\pdfoutput=0
\usepackage{hyperref}
\usepackage[utf8]{inputenc}
\usepackage{graphicx}
\usepackage{amssymb}
\usepackage{amsmath}
\usepackage{graphicx}
\newcommand{\boldm}[1] {\mathversion{bold}#1\mathversion{normal}}
\usepackage[ruled,vlined,linesnumbered]{algorithm2e}
\usepackage{subfig}
\usepackage[skip=4pt]{caption}

\usepackage{algpseudocode}
\usepackage{etoolbox}
\usepackage{booktabs}

\begin{document}
\title{Distance-Guided GA-Based Approach to Distributed Data-Intensive Web Service Composition}
\titlerunning{Distance-Guided DWSC}
%
%
\author{Soheila Sadeghiram \and
	Hui Ma \and
	Gang Chen}
\authorrunning{S. Sadeghiram et al.}
%
\institute{School of Engineering and Computer Science, Victoria University of Wellington\\ New Zealand \\
	\email{\{Soheila.Sadeghiram|Hui.Ma|Aaron.Chen\}@ecs.vuw.ac.nz}}

\maketitle              
\begin{abstract}
Distributed computing which uses Web services as fundamental elements, enables high-speed development of software applications through composing many interoperating, distributed, re-usable, and autonomous services. As a fundamental challenge for service developers, service composition must fulfil functional requirements and optimise \textit{Quality of Service (QoS)} attributes, simultaneously. On the other hand, huge amounts of data have been created by advances in technologies, which may be exchanged between services. Data-intensive Web services are of great interest to implement data-intensive processes. However, current approaches to Web service composition have omitted either the effect of data, or the distribution of services. \textit{Evolutionary Computing (EC)} techniques allow for the creation of compositions that meet all the above factors. In this paper, we will develop \textit{Genetic Algorithm (GA)}-based approach for solving the problem of \textit{distributed data-intensive Web service composition (DWSC)}. In particular, we will introduce two new heuristics, i.e. \textit{LCS (Longest Common Subsequence)} and \textit{distance of services}, in designing crossover operators. Additionally, a new local search technique incorporating \textit{distance of services} will be proposed. 

\end{abstract}
\section{Introduction}
Web services are software modules accessible by other programs over the Web to accomplish a task. They have the ability to provide easy composition of distributed applications in heterogeneous environments \cite{channabasavaiah2003migrating}. Web services, which are provided by service providers and distributed over different locations, require some inputs, and subsequently generate a set of outputs after execution. However, in most cases, individual Web services are composed together to create new services which are able to achieve some complex goals successfully, e.g provide new functionality. As many Web services deliver the functionality, non-functional properties i.e. Quality of Service (QoS), such as response time and cost, are necessary for combining appropriate services. 

Additionally, the size of data available on the Web, is duplicating every 18 months \cite{gantz2009economy}, and many companies need to perform large scale data analysis. Data-intensive Web services deal with huge amounts of data, distributed in different locations in the network, and produce a huge amount of output data. Therefore, a data-intensive Web service composition (DWSC) is defined as the task of selecting data-intensive services from different locations and composing them together to optimise QoS requirements, where the overall quality is ensured by the careful transmission of mass data. Accordingly, the challenges of DWSC is that the quality of composite service are affected by not only the QoS of components, but also by the location of Web services and the size of data transferred between services.

The WSC problem can be explained as a global optimisation problem \cite{alrifai2008distributed}. Finding an appropriate solution by using services from a large repository of Web services can be very difficult and time consuming, because there will be a huge number of possible solutions to choose from in a limited time. Moreover, there are typically two tasks to be considered in composing Web services in a fully-automated manner: preserving the functional correctness of solutions, i.e. ensuring all the services inputs can be fulfilled by the service task or by its proceeding services, and optimising solutions according to their overall QoS. Such problems are known to be NP-hard \cite{song2006dynamic}. 

Therefore, DWSC as an NP-hard combinatorial optimisation problem (COP) cannot be solved efficiently with traditional methods such as mixed linear programming solvers and dynamic programming \cite{bellman2013dynamic}. High computational complexity in solving NP-hard COPs makes these algorithms inappropriate for WSC problems. In this case, computing an optimal solution is impractical and researchers are more interested in building efficient and effective approaches for finding near-optimal solutions.

On the other hand, Evolutionary Computing (EC) techniques have shown to be more efficient than traditional methods for solving difficult problems and, in particular, for the WSC problem when the task increases in complexity \cite{aversano2006genetic}. EC algorithms are population-based search techniques with the ability to search different areas of possible good solutions.

In a Genetic Algorithm (GA), however, the global random search will cause a decline in results. Therefore, applying local search techniques to better balance between exploration and exploitation seems to be useful. Accordingly, there is an enormous effort to consider hybrid versions of EC techniques that incorporate local search techniques resulting in various Memetic Algorithms (MAs) [16]. MAs are often reported to achieve a clearly better performance [18, 11] for solving WSC problems.

In this paper, to solve the DWSC problem, we will design a GA-based approach which considers the nature of the distributed data-intensive services. In particular, we design powerful crossover operators for GA which incorporate the domain knowledge of DWSC to diversify populations, and transfer good building blocks of parents to offsprings. Two heuristics will be used: distance of services and longest common subsequence of parents.

Motivated by the large-scale DWSC problem, the aim of this paper is to propose an effective fully-automated GA-based method for solving the DWSC problem, considering the distributed nature of data-intensive services. To this end, we will consider the distributed environment as in \cite{sadeghiram2018cluster}, which considers location of services and the communication delay and cost (between two services, for example) to improve end-to-end QoS. In this paper, first three different problem-specific crossover operators will be developed through applying distance and longest common subsequence heuristics. Second, a local search for the GA will be proposed for the DWSC problem. Our crossover operators will be compared to the crossover operator suggested in \cite{da2018evolutionary} for the WSC problem, and to each other, as well.

We will propose an effective GA-based approach for DWSC. In particular:
\begin{itemize}

	\item We will propose three different crossover operators which use heuristics to preserve valuable information through generations and to apply domain knowledge; 
	\item We will propose a local search technique for the DWSC problem; 
	\item We will compare the performance of these different crossover operators, and a baseline crossover operator for WSC. 
 
\end{itemize}
The rest of this paper is organised as follows: Section \ref{related} presents related works. Section \ref{defination} presents the problem definition and the quality model of distributed services. Section \ref{design} describes the algorithm design. Section \ref{ev} describes experimental evaluation results and discussions of the results. Finally, Section \ref{conclusion} concludes the paper. 
\section{Related Works}\label{related}

Various EC techniques have been applied to solve the WSC problem in the literature, examples are GA for WSC \cite{da2018evolutionary,da2016memetic,canfora2005approach}, Genetic Programming (GP)\cite{koza1992genetic} for WSC \cite{da2016memetic,da2018evolutionary} and Particle Swarm Optimisation (PSO) \cite{da2016particle}. The main difference between GA and GP is that GA uses sequences to represent chromosomes (solutions), while in GP, solutions are represented as trees. The application of genetic operators on sequences in GA is more straightforward than trees in GP, i.e., operators can be applied without restrictions since the functional correctness of the composition will be ensured during decoding, which will be explained in more detail in section \ref{design}. Additionally, the superiority of the GA-based method to Particle Swarm Optimisation \cite{kennedy2011particle} and integer linear programming methods are demonstrated in \cite{da2018evolutionary} and \cite{canfora2005approach}, respectively.

Existing WSC approaches assume a centralised service repository for the composition. Conversely, in a DWSC of great importance are the size of data, and the location of a Web service which determines its distances to other Web services and the communication cost and time.
Additionally, a majority of existing methods for solving DWSC are semi-automated \cite{bousrih2015optimizing,wang2016multi,mohsni2016data}. As opposed to fully-automated approaches, they depend on a predefined abstract process model, i.e., a predefined workflow which embodies a set of tasks and their data dependency. Clearly, these workflows should be statically generated by the service requester or at the design time. The WSC process is limited to only selecting and binding single Web services to each task in the workflow rather than constructing relevant workflows.
Furthermore, none of them considers the communication between services in the network.
 For example, the approach for DWSC in \cite{bousrih2015optimizing} only tackles selecting Web services for each slot in a predefined workflow and applies a data placement strategy using a dependency matrix of data in order to reduce the transfer cost \cite{bousrih2015optimizing}. Data items which are mostly used together, are grouped and placed on the same server before the composition phase starts. Moreover, in the above approach, services are stored in the same location as data they use, in order to lower the frequency of data, and for diminishing the cost and response time of composite services. Though using the dependency information increases the effectiveness of the method, this idea does not apply in real application because placing data and services on data centres should be supported separately by data providers. 

Another group of approaches have investigated fully-automated DWSC \cite{yu2015f,yu2014hybrid,sadeghiram2018cluster}. A hybrid approach combining GP and Tabu Search \cite{glover1989tabu} is proposed to solve the DWSC problem \cite{yu2014hybrid}. However, this method does not clarify how it checks the validity of a solution. Another approach to DWSC is proposed in \cite{yu2015f}, in which tree-based representation and search space reduction techniques are adopted before the optimisation starts using fully and partially dependent Web services. In another paper, a clustering-based GA algorithm which uses sequence-like representation for fixed-length GA chromosomes has been suggested \cite{sadeghiram2018cluster}, where the nature of Web service's distribution is incorporated in generating the initial population of GA. 

In this paper, we propose a fully-automated DWSC approach which takes in to account the distribution of services and the communication.

\section{Problem Definition and Objective Function}\label{defination}

The performance of a distributed DWSC is affected by the data transferred between services at different locations. In fact, the challenge of a distributed DWSC is that selecting component services for service composition demands to consider QoS and relative locations of Web services to each other for transfering input and output data, because all these factors have a significant impact on the overall QoS of service composition. 

 In this section, we will first present the definition of the distributed DWSC based on the definition in our previous paper \cite{sadeghiram2018cluster}. We will present some basic terms for the distributed DWSC problem, and then will present the objective function of the problem.

\subsection{Basic Terms}\label{problemDes}
First, we define basic concepts that need to be used later in the objective function. We consider a \textit{data-intensive Web service} as a tuple \textit{ $S_{i}$= ($I_{i}$, $O_{i}$, $QoS_{i}$, $D_{i}$, $l_{i}$ )} where {\textit{$I_{i}$}} is a set of inputs. $O_{i}$ is a set of outputs of the service. $QoS_{i}$ is the set of quality attributes of the service which describes non-functional properties. In this paper, for each Web service we define $T_{i}$ and $C_{i}$, which refer to the total time and cost required for executing that service. $D_{i}$ is the set of $m$ data items $d_{j}$, $ j \in \{1,...,m\}$ required by the service, and $l_{i}$ is the location of the Web service.

A \textit{service repository} $\mathcal{R}$ consists of a finite collection of Web services $S_{i}$, $ i \in \{1,...,n\}$ together with their input/output specifications. A \textit{service request} (also called a \textit{composition task}) is a tuple $\mathcal{T}=(I_\mathcal{T}, O_\mathcal{T})$ where $I_\mathcal{T}$ is a set of task inputs a customer can provide for the composition, and $O_\mathcal{T}$ is a set of task outputs expected by the user to be produced by the composition.

\textit{Data} is defined as a tuple \textit{ d= ($cost_{d}$, $size_{d}$, $l_{d}$ )}
where \textit{$cost_{d}$} is the cost applied by the data provider, i.e. the cost to provide data, \textit{$size_{d}$} is the data size, and \textit{$l_{d}$} is the location of the data centre that hosts the data.

Results of the service composition request are often represented as Directed Acyclic Graphs (DAGs) which include a set of services that could jointly accomplish the required task, where two special services can be used to represent the overall composition's inputs and outputs: a start service $S_0$ with $I(S_0)$ = $\varnothing$ and $O(S_0)$ = $I_\mathcal{T}$, and an end service $S_{n+1}$ with $I(S_{n+1})=O_\mathcal{T}$ and $O(S_{n+1})$= $\varnothing$. In a composite Web service, there is a communication link between $S$ and $S'$ if some outputs of $S$ serve as inputs for $S'$. We need to ensure that composite services are $feasible$. In particular, we need to ensure that any two directly connected services in the DAG match considering their inputs/outputs. 

In this paper we have made the following general assumptions for the DWSC problem: first, all data and Web services have been deployed on servers across the network before the composition starts; second, the required data by each service should be transferred to the location of that service before its execution.

\subsection{Quality of Composite Services}
During the execution of a composed data-intensive Web service in a distributed environment, a large amount of time is needed to transfer and access data. The following definitions have been used in this work to accurately capture the various time and cost components involved \cite{sadeghiram2018cluster}:
\begin{itemize}
	\item Server access latency \begin{math}({Tsal})\end{math}: the storage (server) related time, which depends on the server attributes and the size of data to be retrieved.
	
	\item Data process time \begin{math} ({Tproc})\end{math}: the service related time for processing a data.
	
	\item Service cost \begin{math} ({Cs})\end{math}: the cost to process the data and the cost to provide the service.
	
	\item Data provision cost \begin{math} ({Cprov})\end{math}: the cost applied by data provider, i.e., the cost to provide the data.
	\item Data transfer time \begin{math} ({Tt})\end{math}: the amount of time that it takes for a server to put all bits of the given data over the communication link and is defined in terms of the data size over the network bandwidth.
\end{itemize}

The distributed nature of DWSC is further captured through two quantities:
\begin{enumerate}
	
	\item Propagation delay \begin{math} ({Tp})\end{math}: the duration that it takes for the data to travel from the sender to the receiver over a given path. The sender is either a server hosting the data or a server hosting a preceding service in the composition which produces required data for the receiver. In the first case, we denote the propagation time as \begin{math} {Tpd}\end{math} while in the second case we denote it as \begin{math} {Tps}\end{math}. Similarly, the receiver is a server hosting a Web service which takes that data as its input. 
	Propagation delay can be computed as the ratio between the link length and the propagation speed over the specific medium (the wave propagation speed). 
	In this paper, we define the propagation delay in terms of the distance between two servers.
	\item Communication cost \begin{math} ({Cc})\end{math}: the cost to transfer a data from one Web service to another. We denote it as \begin{math} {Ccs}\end{math} if the data is transferred between two communicating Web services, while we use \begin{math} {Ccd}\end{math}  to demonstrate the cost to move data from the hosting server to a Web service. 
\end{enumerate}

Accordingly, $T_{i}$ and $C_{i}$ (the total execution time and cost of a Web service $S_{i}$ including data-related time and cost) are calculated in Equations \eqref{tatomic} and \eqref{catomic}, respectively.

\begin{equation}
T_{i}={\sum_{j=1}^{m}(Tpd_{d_{j}}+Tsal_{d_{j}}+Tproc_{d_{j}}+Tt_{d_{j}})}
\label{tatomic}
\end{equation}

\begin{equation}
C_{i}={\sum_{j=1}^{m}(Ccd_{d_{j}}+Cd_{d_{j}}+C_s})
\label{catomic}
\end{equation}

In the above functions, $m$ is the total number of data items in $D_{i}$.

The overall cost is obtained by adding together individual values of single services in the composition, i.e., nodes and associated costs for edges in the graph, and it is calculated in the same way for both parallel and sequence constructs in Equation \eqref{ci}:

\begin{equation}
C_{total}=w_{c1}\sum_{i=1}^{NODE} C_{i}+w_{c2}\sum_{i=1}^{EDGE}{Ccs_{i}}
\label{ci}
\end{equation}
\noindent \begin{math} {Ccs} 
\end{math} is the communication cost, and \begin{math}
w_{c1} \end{math} and \begin{math} w_{c2} \end{math} are positive real values to assign weights to each term. These parameters indicate the importance of each corresponding attribute. \begin{math} NODE \end{math} and \begin{math} EDGE \end{math} are the total number of nodes (Web services) and edges (links between services) included in that composition, respectively. 

Response time $T_{total}$ is the time of the most time-consuming path in the composition, i.e., assuming $h$ is the number of path,
 \begin{math} N{j} \end{math} and \begin{math} E{j} \end{math} are the number of nodes and the number of edges in a path, respectively. The overall time is defined as in Equation \eqref{ti}:  
\begin{equation}
T_{total}=MAX\{({w_{t1}}\sum_{i=1}^{N{j}}{T_{i}}+w_{t2}\sum_{i=1}^{E{j}}{Tps_{i}})|j\in\{1,2,..,h\}\}
\label{ti}
\end{equation}

 \begin{math}w_{t1} \end{math} and \begin{math} w_{t2} \end{math} are positive real weights.

Finally, we use the objective function in Equation \eqref{fitness} to evaluate QoS of DWSC solutions. The best solution will be a composition with the minimum objective value. This function will be used as fitness in our EC algorithm. 
\begin{equation}
ObjectiveFunction=\hat{T}_{total}+\hat{C}_{total}
\label {fitness}
\end{equation}
where \begin{math}
\hat{T}_{total}\end{math} and \begin{math}\hat{C}_{total} \end{math} are normalised values of \begin{math}{T}_{total}\end{math} and \begin{math}{C}_{total} \end{math}, respectively. The upper bound for normalisation is calculated as it is discussed in \cite{da2018evolutionary}.

\section{GA-Based Algorithm for DWSC}\label{design}

GA maintains a population of individuals and uses the current population for defining the promising region \cite{holland1992genetic}. It relies on crossover to combine parts of different individuals, to derive new solutions. New individuals may be modified by other operators, such as mutation, before being added to population. The pseudocode of the GA-based method for DWSC is shown in Algorithm \ref{algorithm1}. Distance-guided crossover operator is one of the new parts of the algorithm proposed in this paper. One of the four crossover operators, which we discuss in this paper, will be applied. In addition, a novel local search is developed and used.

\begin{algorithm}[!htb]
	\setlength\hsize{0.9\linewidth}
	\SetKwInOut{Input}{Input}\SetKwInOut{Output}{Output}
	\SetKwFunction{filterByLayer}{filterByLayer}\SetKwFunction{mergeLayers}{mergeLayers}\SetKwFunction{findHighestTime}{findHighestTime}
	\SetKwFunction{eNull}{null}\SetKwFunction{getInputsSatisfied}{getInputsSatisfied}\SetKwFunction{calculateFitness}{calculateFitness}
	\LinesNumbered
	\SetNlSty{}{}{:}
	\Input{$Service~Repository$}
	\Output{$A~Service~Composition~Solution$ }
	
		Generate sequences with randomly ordered services in the $service repository$;
		
		Apply Decoding to create a graph for each sequence and Calculate fitness of each sequence;
		
		Update sequences by removing redundant services;

		\While{number of iterations not reached}{
			Select Individuals based on their fitness value using a tournament selection;
			
			Apply distance-guided crossover operator;
			 
			Apply mutation operator;
			
			Apply local search operator;
			
			Apply Decoding and calculate $fitness$ for each graph;
			\If{fitness is better than Bestfitness}{
				$Bestfitness$=$fitness$;
	}}
	
	\Return $SequenceWithBestFitness$\;
	\caption{Proposed algorithm for DWSC}
	\label{algorithm1}
\end{algorithm}

In this paper, we use sequence-like representation to reference composition solutions. This representation was firstly adopted for the WSC problem in \cite{da2016particle} and then successfully applied to other works in this domain \cite{sadeghiram2018cluster,da2016memetic,da2018evolutionary}. 
The initial population is created by randomly ordering all the services in the repository, in sequences. GA operators and local search are applied on sequences.

 In this paper, we adopt variable-length problem-specific crossover and mutation operators, as well as a local search which has been specifically designed for the DWSC. In particular, heuristics are used in defining new crossover operators for the DWSC. Distance of services, as a heuristic contributes to all three crossover operators and to the local search. Considering distance in distributed DWSC has shown to be effective, as a heuristic in creating initial population for GA with fixed-length chromosomes.

 The mutation follows the same idea as the mutation operator for WSC research \cite{da2018evolutionary}, without any changes. 
 Distance-guided crossover operators and local search will be discussed in \ref{operators} and \ref{loc}, respectively.

\subsection{Representation of Chromosomes}
 For the WSC problem, different representations have been investigated in \cite{da2018evolutionary}, but their core idea is to generate candidate services compositions encoded as sequences of service. Sequences are converted to executable compositions, i.e. $feasible$ workflows, based on services input/output through decoding. A backward decoding algorithm is then used to transform any sequences of services into a corresponding feasible service composition, i.e. a feasible workflow \cite{da2018evolutionary,sadeghiram2018cluster}. An example of backward decoding of a sequence, and the sequence after being decoded are illustrated in Fig.1. In fact, the decoding scheme generates workflows automatically which is the requirement of a fully-automated DWSC method. Resulting workflows cover parallel and sequence constructs, which frequently appear in a composed Web service. Our EC operators will produce chromosomes with duplicated services that affect the efficiency of the algorithm. Therefore, duplicated services are removed from the sequence before the decoding starts. Moreover, as it is illustrated in Fig.1(a), during the decoding, each candidate is reduced to contain only those services used in the composition solution. For more information about the decoding algorithm refer to the backward decoding in \cite{da2018evolutionary,da2016particle}. Additionally, sequences must include enough services so that decoding could produce feasible solutions.
 
 \begin{figure}
 	
 	\label{fig:backdecode}

 	 \centering
 	\begin{tabular}{@{}c@{}}
 			\includegraphics[width=9cm]{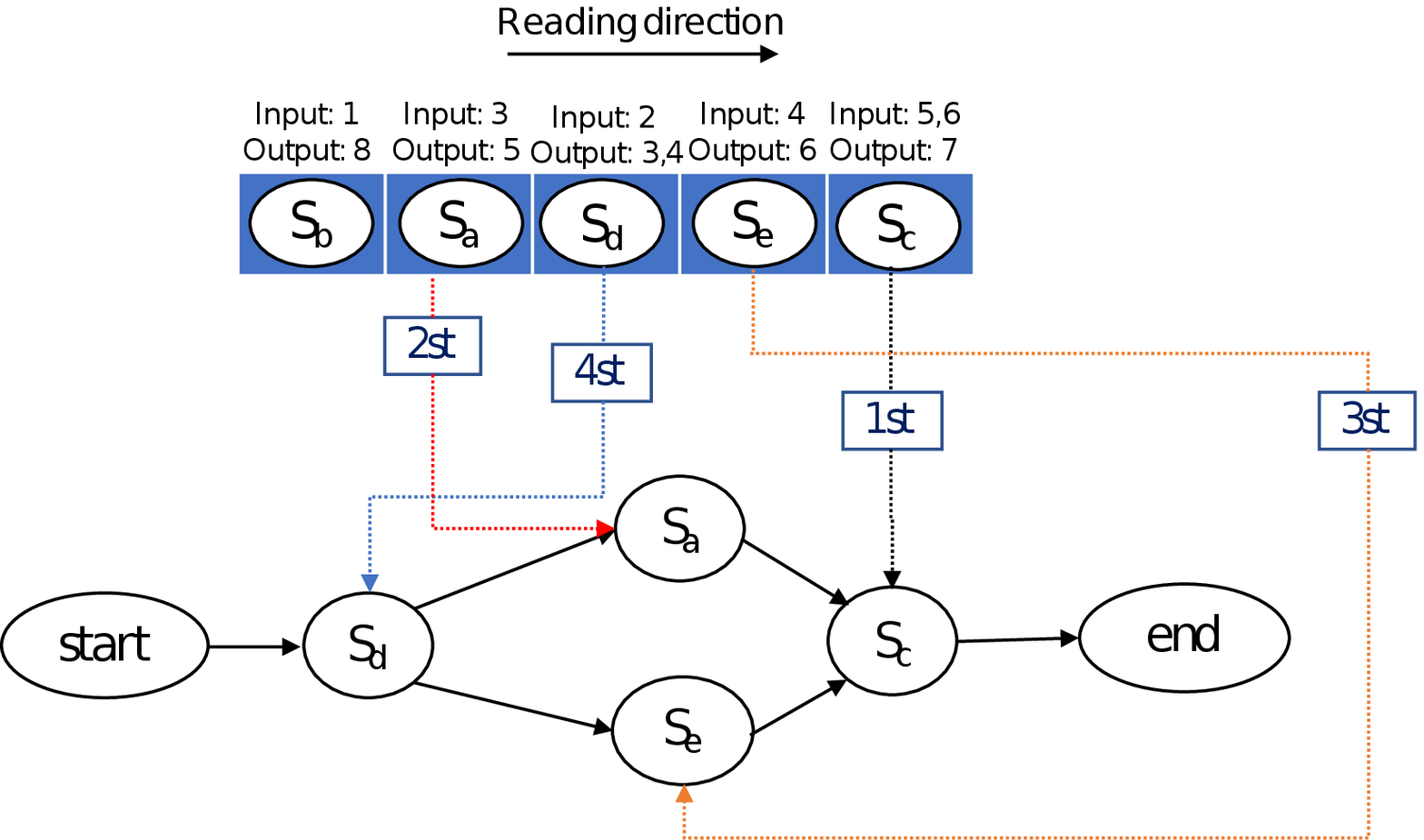}\\[\abovecaptionskip]
 		\label{fig:backward}
 		

 		\small  (a) 
 	\end{tabular}
 	
 	
 	\begin{tabular}{@{}c@{}}
 		\includegraphics[width=4cm]{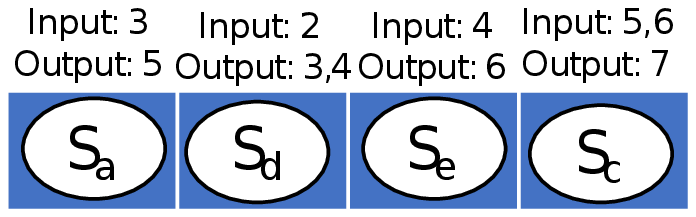}\\[\abovecaptionskip]
 	\captionsetup{justification=centering}
 	\label{fig:afterdecod}

	\small (b) 
 	\end{tabular}
 	
 	
 	\caption{(a) Backward decoding example (note that the sequence is traversed as many times as possible), (b) the sequence after being decoded in variable-length GA.}
 	
 \end{figure}

\subsection{Distance-Guided Crossover Operators for the DWSC}
 \label{operators}
\begin{figure}
		\centering
	\includegraphics[width=0.8\textwidth]{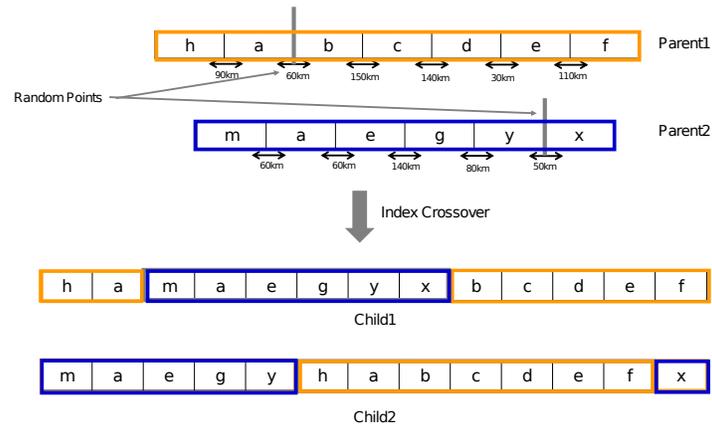}
	\caption{Example of index crossover}
	\label{fig:indexcrossover}
\end{figure}

We will use a crossover operator previously applied to WSC for variable-length chromosomes as a baseline method. Existing research on WSC using problem-specific variable-length chromosomes, determines a random crossover point for each parent \cite{da2018evolutionary}, where indices are independently chosen for the two parents. Then, each parent is split from the index point into a $prefix$ and a $suffix$. In order to generate offspring, each original parent is embedded within the $prefix$ and the $suffix$ of the other parent. The functionality of this crossover is demonstrated in Fig. \ref{fig:indexcrossover}. This crossover is designed specifically for the problem of WSC and no domain knowledge is used. We will use this crossover to obtain three other crossovers for our problem, step-by-step, which use distance information, and we call it \textit{index crossover}. This crossover maintains feasibility, because its offspring inherits all the services from their parents.

In this work, new variable-length problem-specific crossover operators are proposed to add diversity, maintain feasibility and incorporate the domain knowledge. The effect of three different factors in designing such a suitable crossover has been investigated: the longest common sub-sequence (LCS) of services within two individuals, geographical distance between each pair of services, and the amount of diversity that the crossover generates. In addition, it is necessary to define these operators in a way that no infeasible individuals are generated.

\begin{figure}
			\centering
	\includegraphics[width=0.8\textwidth]{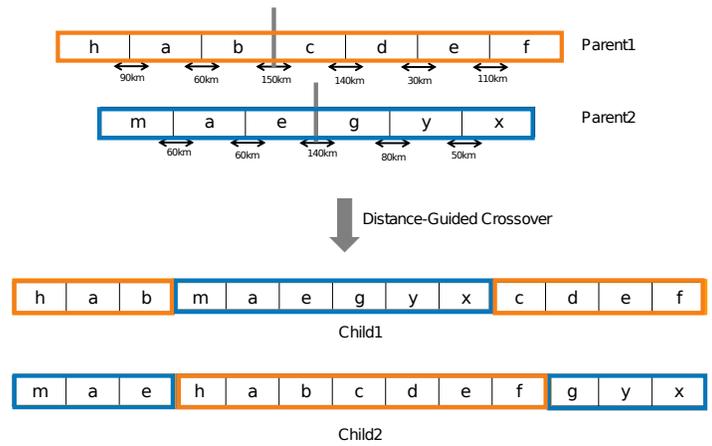}
	\caption{Example of distance-guided crossover operator}
	\label{fig:distance-guided crossover}
\end{figure}

\paragraph {Distance-guided index crossover:} this crossover is designed to incorporate domain knowledge (e.g. location of services) into GA. Location of services are applied as heuristics into the crossover operator. As illustrated in Fig.\ref{fig:distance-guided crossover}, this crossover is very similar to \textit{index crossover} mentioned above, except that instead of choosing crossover points in random, a crossover point is selected based on the distance of services to each other in the sequence. In other words, in a sequence of services, the crossover point is selected in a way that the distance between the two consecutive services adjacent to this point is greater than the distance between any other pairs of consecutive services. In fact, this new crossover operator combines index crossover in the distributed DWSC problem with the domain knowledge, i.e., through the location of services. 


\paragraph {Distance-guided two-point crossover:} the difference between this crossover and the two previously described crossover operators (i.e., index crossover and distance-guided index crossover), is that this is a two-point crossover rather than a single point one, where indices in each parent are chosen based on the first and second greatest distance between pairs of consecutive services. Therefore, using these two indices, each parent is split into three parts. Afterwards, to create children, the three parts of the first parent are combined with the three parts of the second parent, one in between. Two offspring differ each other in the order of combination. The aim of this crossover is to increase diversity in the resulted sequences through breaking parents apart into three parts. In other words, though this crossover is very similar to the \textit{distance-guided index crossover} in that both embed domain knowledge, the diversity is increased between children and parents because each parent is split into three parts rather than only two parts. 

\begin{figure}
	\centering
	\includegraphics[width=0.9\textwidth]{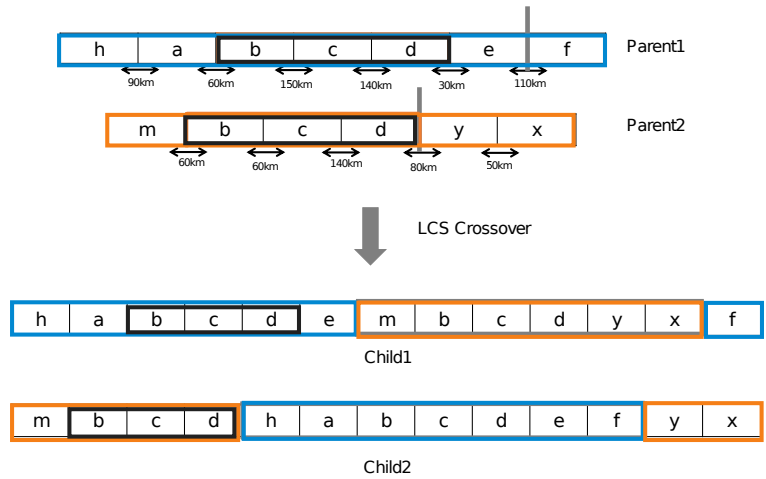}
	\caption{Example of distance-guided LCS crossover for the proposed method. The longest common subsequence is identified by orange rectangle.}
	\label{fig:LCS crossover}
\end{figure}

\paragraph {Distance-guided LCS crossover:} in this crossover, which is indicated in Fig.\ref{fig:LCS crossover}, a heuristic is incorporated to preserve the promising part of each parent and transfer it to children. This new heuristic is called longest common subsequence (LCS) between two parents, i.e. the longest sequence of services which appears in both parents. The LCS is repeated in the offspring without any changes. For this purpose, the crossover point is selected in the same way as the \textit{distance-guided index crossover} except that this point cannot be inside the LCS. In that way, the crossover preserves the LCS from being split  which can be a part of a good solution. The feasibility is maintained in the same way as it was conducted in above crossovers. Duplicated services as well as unused services in the composition will be removed through decoding.

\subsection{Distance-Guided Local Search for the DWSC}\label{loc}
\begin{figure}
	\centering
	\includegraphics[width=0.8\textwidth]{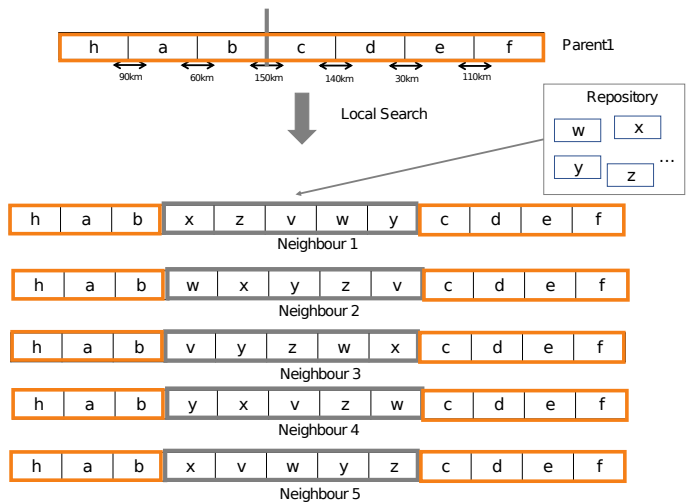}
	\caption{Example of local search and creating five different neighbours.}
	\label{fig:local search}
\end{figure}

In a GA, however, the global random search will cause a decline in results. Therefore, applying local search techniques to better  balance between exploration and exploitation seems to be useful. Accordingly, there is an enormous effort to consider hybrid versions of EC techniques that incorporate local search techniques resulting in various Memetic Algorithms (MAs) \cite{moscato2004memetic}.
MAs are often reported to achieve a clearly better performance \cite{da2016memetic,jula2013hybrid} for solving WSC problems. 
In this paper, a local search operator creates its neighbourhood by repeatedly inserting a group of Web services into the original individual (see Fig.\ref{fig:local search}). Once more, the idea of distance is utilised for selecting a point in the individual where the new sequence of services will be inserted. For example, if the distance between the two consecutive services $a$ and $b$ is the largest, all services from the repository whose outputs can be used to fulfil $b$ are inserted in between $a$ and $b$. Each neighbour differs from others in terms of the order of services in the newly added sequence. 

\section{Evaluation}\label{ev}
 The aim of this paper is to propose a GA-based approach to the problem of distributed data-intensive service composition in a fully automated manner. In this section, we will conduct experimental evaluation to evaluate our proposed GA method.

\subsection{Experiment Design}\label{ex}

A set of experiments is carried out using WSC-2008 \cite{bansal2008wsc} and WSC-2009 \cite{kona2009wsc} benchmark datasets. WSC-2008 contains eight service repositories of varying sizes, while WSC-2009 has five repositories with a greater variety of sizes. A taxonomy of concepts is provided for determining which inputs and outputs are compatible, and a number of service composition tasks are also given \cite{bansal2008wsc,kona2009wsc}. These datasets were chosen because they are the largest benchmarks that have been broadly exploit in the composition literature; however, datasets do not contain the location information of the server which hosts Web services nor the data-intensive information. The distance between two Web services is estimated using the same method as proposed in the previous paper \cite{sadeghiram2018cluster} based on the information in WS-dream open dataset \cite{zheng2014investigating}.

Random values are assigned to \begin{math} {Tt}\end{math} for each connection line in the interval (0,1], as well as to \begin{math} {Ccs}\end{math} and \begin{math} {Ccd}\end{math}. 
Additionally, each server has its own \begin{math} {Tsal}\end{math} and each data has its own \begin{math} {Ccd} \end{math} similarly, which are both generated randomly in that interval.

A data item provides each service with a price and response time which we assigned them to \begin{math} {Tproc}\end{math} and \begin{math} {Cs}\end{math}, respectively. These values are  normalised to fit between 0 and 1.
Values of weight parameters are \begin{math} w_{t1}=w_{c1}=0.2 \end{math}, and \begin{math} w_{t2}=w_{c2}=0.3 \end{math}. Accordingly, \begin{math} w_{c1}+w_{c2}=0.5 \end{math}, and \begin{math} w_{t1}+w_{t2}=0.5 \end{math} which means that the time and cost have identical contributions to the fitness function. Finally, the sum of all weights is equal to one, \begin{math} w_{t1}+w_{t2}+w_{c1}+w_{c2}=1 \end{math}, which means that the final fitness value will fit between 0 and 1. 

Four methods are compared, i.e. variable-length GA-based method with four different crossovers, each of them being run 30 independent times on desktop computers with 8 GB RAM and an Intel Core i7-4790 processor (3.6GHz). All methods have the same parameter set.  Probability of local search, mutation and crossover are 0.05, 0.05 and 0.95, respectively. Neighbourhood size for local search is 10. Tournament selection with the size of 2 is adopted as the strategy for choosing which candidates to update in these approaches. For example, individuals to undergo local search are chosen through a tournament selection of size 2. Elitism size is 2 for all methods. To evaluate our proposed distance-guided GA-based method, and to make a fair comparison, all parameters have been set equally. All proposed methods use the same local search technique. 

\subsection{Results}

 Table \ref{tab:results} shows the mean solution fitness and standard deviation for 30 independent runs of each approach. Distance-guided GA-I and GA-II indicate distance-guided index and two-point approaches, respectively. Wilcoxon signed-rank tests at 0.05 significance level have been carried out to verify whether fitness values were significantly different. Pairwise comparisons have been conducted for all possible combinations of methods on all datasets. The comparison results then have been used to rank each approach and identify the top performers. For example, for dataset 08-2 in Table\ref{tab:results}, the fitness achieved by the GAII was significantly better in all pairwise comparisons against other approaches. On the other hand, for dataset 08-1 there was no significant difference between GA-LCS and GA-II. 
Results show that quality of solutions which are produced using distance-guided crossover operators, i.e., GA-I, GA-II and GA-LCS, are generally better than results of index crossover. The reason can be explained in the shortcoming of index crossover, which has been adopted from a solution to the WSC problem \cite{da2018evolutionary} and is not specifically designed for the distributed DWSC problem. Moreover, GA-LCS have produced the best results out of the four crossovers, presumably due to maintaining useful subsequence without breaking them into separate parts. In addition, considering GA-II and GA-I, the former has indicated better results. This means that increasing the difference between offspring chromosomes and their parents improves the quality of solutions. Finally, results in Table \ref{tab:results} confirms that a more powerful crossover operator for GA, which uses heuristics and the domain knowledge, can produce better results than those of a general method. 
\begin{table}
	\centering
	\caption{Mean fitness values and standard deviations per 30 runs. Significantly better values in all four comparisons are displayed in bold. (Note: the lower the fitness the better)}
	\label{tab:results}
	\begin{center}
		
		\newcommand{\abcdresults}[6]{ #1 & #2 & #4 &#5 & #6\\}
		\begin{tabular}{|l|l|l|l|l|}

			\hline
			
			\multicolumn{1}{|p{2cm}|}{\centering Set of Experiments} &  \multicolumn{1}{|p{2cm}|}{\centering GA for WSC (Index) \cite{da2018evolutionary} } & \multicolumn{1}{|p{2.5cm}|}{\centering Distance-guided GA-I}  &\multicolumn{1}{|p{2.5cm}|}{\centering Distance-guided GA-II } & \multicolumn{1}{|p{2.5cm}|}{\centering Distance-guided GA-LCS }   \\

			\hline

			WSC08-1  & {\begin{math} 0.54\pm0.04\end{math}}  & {\begin{math} 0.45\pm0.01\end{math}} & {\boldm $ 0.42\pm0.04$}   & {\boldm $0.42\pm0.02$}  \\
			
			WSC08-2  & {\begin{math} 0.51\pm0.09\end{math}} & {\begin{math} 0.46\pm0.03\end{math}} &{\boldm $0.42\pm0.05$} & $ 0.46\pm0.04$ \\
			
			WSC08-3   & {\begin{math} 0.55\pm0.18\end{math}} &{\begin{math} 0.52\pm0.04\end{math}}   & \begin{math} 0.48\pm0.01\end{math}    &  {\boldm $0.47\pm0.02$}  \\
			
			WSC08-4  & {\begin{math} 0.52\pm0.03\end{math}} & {\begin{math} 0.49\pm0.02\end{math}} & \begin{math} 0.45\pm0.06\end{math}     & {\boldm $ 0.41\pm0.01$} \\
			
			WSC08-5  & {\begin{math} 0.55\pm0.08\end{math}} & \begin{math} 0.52\pm0.01\end{math}  &{\boldm $0.49\pm0.01$}      &  {\boldm $ 0.48\pm0.04$}
			\\
			
				WSC08-6 & \begin{math} 0.58\pm0.13\end{math} &\begin{math} 0.58\pm0.01\end{math} &	\begin{math} 0.55\pm0.02\end{math}     &   {\boldm $0.51\pm0.15$} \\
			
			WSC08-7 & \begin{math} 0.57\pm0.01\end{math} & \begin{math} 0.59\pm0.01\end{math}   &\begin{math} 0.58\pm0.04\end{math}      & {\boldm $0.56\pm0.02$}     \\
			
			WSC08-8 & {\begin{math} 0.54\pm0.08\end{math}} & \begin{math} 0.53\pm0.02\end{math} & {\boldm $ 0.47\pm0.09$}      &  {\begin{math} 0.48\pm0.08\end{math}}  \\
			
			WSC09-1 & {\begin{math} 0.59\pm0.03\end{math}} &{\begin{math} 0.59\pm0.03\end{math}}& \begin{math} 0.57\pm0.07\end{math}     &  {{\boldm $ 0.56\pm0.05$}} \\
			
			WSC09-2   &  {\begin{math} 0.56\pm0.01\end{math}}  & \begin{math} 0.50\pm0.002\end{math}   &  {\boldm $0.49\pm0.1$}   & \begin{math} 0.50\pm0.01\end{math}  \\
			
			WSC09-3 &\begin{math} 0.55\pm0.09\end{math} &{\begin{math} 0.52\pm0.06\end{math}}  & \begin{math} 0.52\pm0.09\end{math}  & {\boldm $ 0.52\pm0.001$}  \\
			
			WSC09-4  &\begin{math} 0.55\pm0.09\end{math} & {\begin{math} 0.52\pm0.01\end{math}} & {\boldm $0.51\pm0.17$} &   {\boldm $0.51\pm0.21$}  \\
			WSC09-5 & \begin{math} 0.58\pm0.09\end{math} & $ 0.49\pm0.02$& \begin{math} 0.51\pm0.02\end{math} & {\boldm $0.48\pm0.06$}     \\
			\hline

		\end{tabular}%
	\end{center}
	\label{tab:1}
\end{table}

\subsection{Discussion}\label{discussion}
\begin{figure}
	\centering
	\subfloat[dataset WSC08-5]{\includegraphics[width=0.5\textwidth]{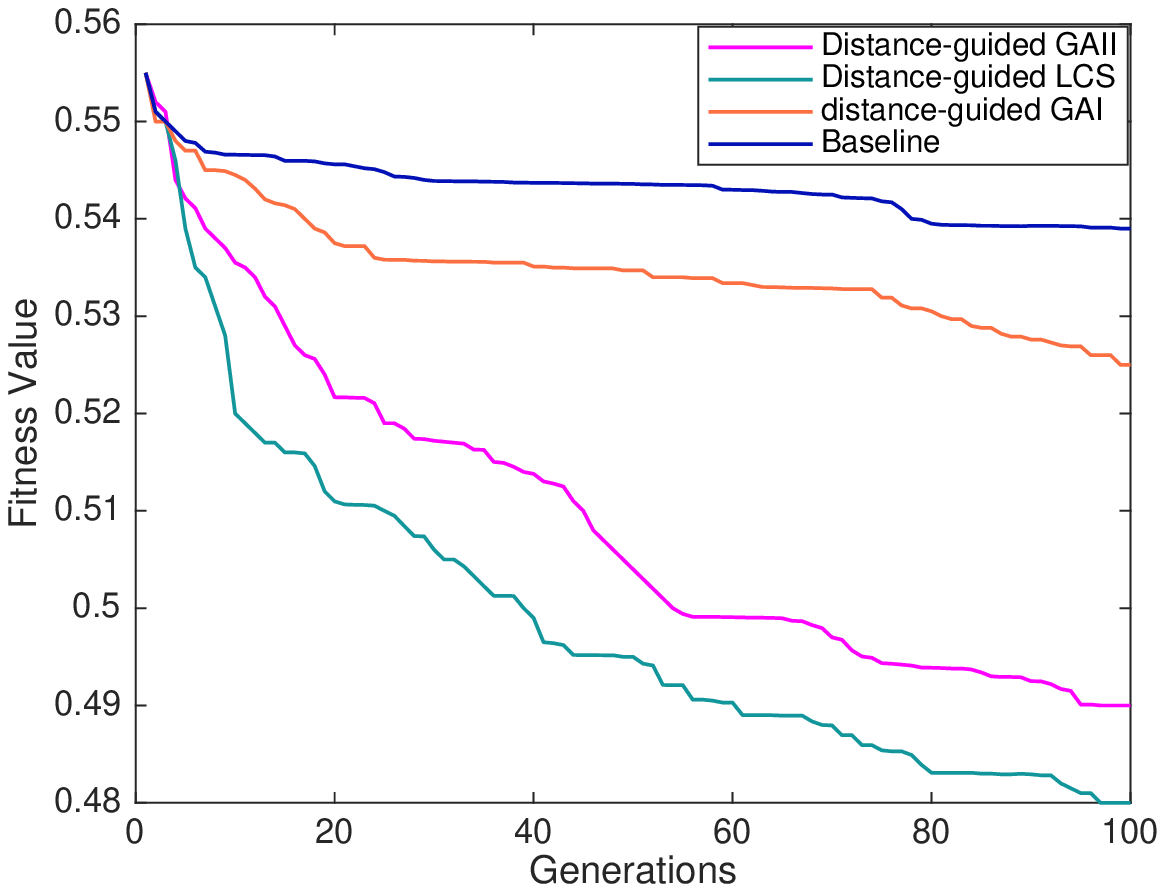} }
	\subfloat[dataset WSC09-4]{\includegraphics[width=0.5\textwidth]{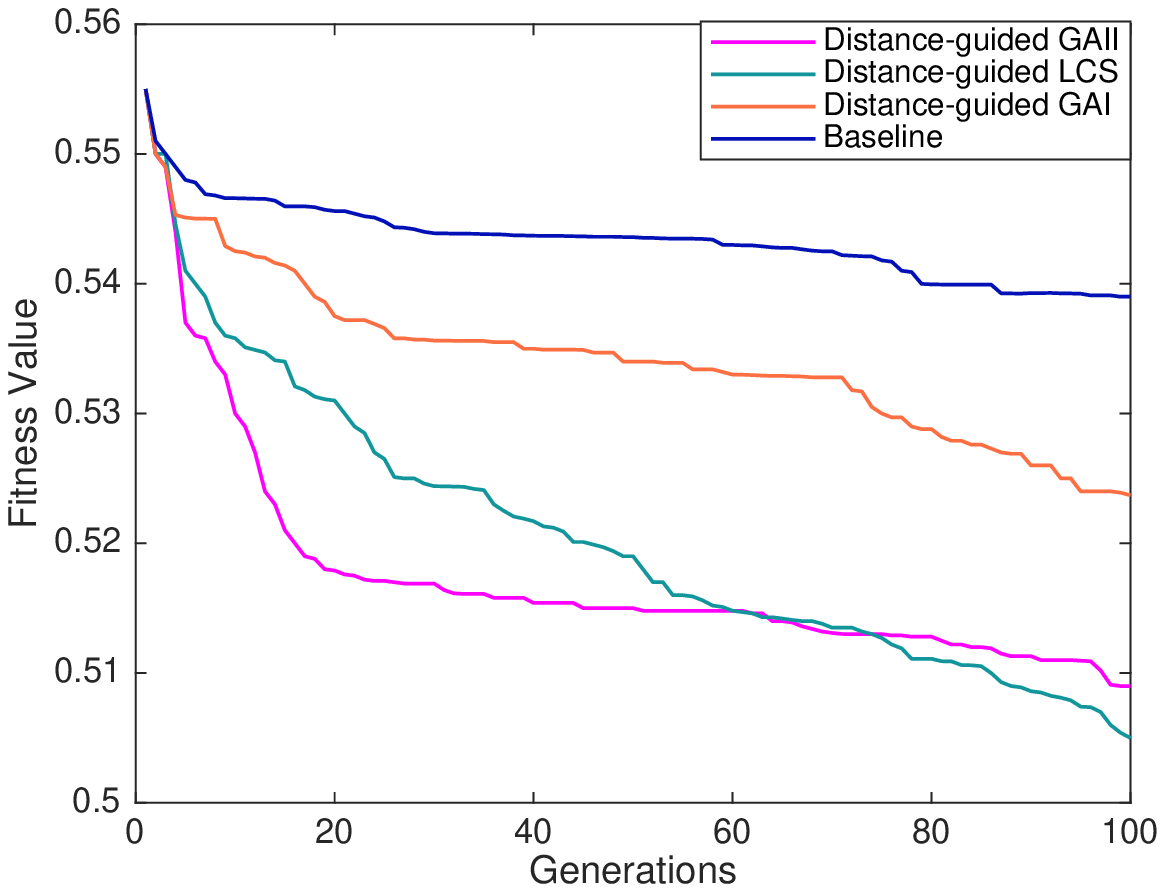}}
	\caption{Mean fitness values over 30 runs.} 
	\label{fig:plot} 
\end{figure}

Fig. \ref{fig:plot}(a) and \ref{fig:plot}(b) illustrate mean fitness values over 30 runs per generations for datasets WSC08-5 and WSC09-4, respectively.
Our experimental evaluation shows that crossovers which utilises distance information of services, i.e. distance-guided crossovers, indicated better fitness value than the other approach, i.e. index crossover. In addition, distance-guided LCS crossover was designed, which maintains promising parts of both parents, adds diversity to compensate for the offspring when it is not very different from its parents, and incorporates domain knowledge, which is, considering the distribution and location of services. These attributes leads to a good performance of the distance-guided LCS method. 
In the future, we will investigate the effect of different local search operators in combination with GA, and inspect where and when to apply them during the search. In addition, designing powerful operators specific for the problem of DWSC can be extended to fixed-length GA, though more efforts are required to set up such operators for GA with fixed-length chromosomes, as the chromosome size is not allowed to be changed.

\section{Conclusion}\label{conclusion}

In this paper we studied the problem of distributed DWSC. We applied an appropriate fitness function to include quality of services, properties of data items used by those services, and communication attributes between services. We proposed a GA-based approach with specific crossover operators and a local search technique. Our experimental evaluation using benchmark datasets shows that our proposed distance-guided two-point GA approach to DWSC can effectively generates better results comparing with other approaches.
These operators help the offspring to inherit good parts from parents. Additionally, distance-guided two-point crossover includes enough diversity to obtain good solutions for DWSC.

%
%
%
%






\end{document}